\documentclass[10pt,twocolumn,letterpaper]{article}

\usepackage{times}
\usepackage{epsfig}
\usepackage{graphicx}
\usepackage{amsmath}
\usepackage{amssymb}
\usepackage{tikz}
\usepackage{pgfplots}
\usepackage{xspace}
\usepackage{bbm}
\usepackage{contour}
\usepackage{colortbl}
\usepackage{multirow}
\usepackage{flushend}

\usepackage[font=small,labelfont=bf,tableposition=top]{caption}
\usepackage{dblfloatfix}

\usepackage[pagebackref=true,breaklinks=true,letterpaper=true,colorlinks,bookmarks=false]{hyperref}

\usepackage{cvpr}
\cvprfinalcopy 
\def\cvprPaperID{758} 

\ifcvprfinal\fi

\usepackage{pgfplots}
\usepackage{pgfplotstable}
\usepackage{pgf,tikz}

\usepgfplotslibrary{external}
\newcommand{\extfig}[2]{\tikzsetnextfilename{figs/extern/#1}{#2}}
\newcommand{\extdata}[1]{\input{#1}}

\newcommand{\leg}[1]{\addlegendentry{#1}}

\newcommand{\comment}[1]{}

\begin{document}
\title{Efficient Diffusion on Region Manifolds:\\
Recovering Small Objects with Compact CNN Representations}

\author{
Ahmet Iscen$^1$ \ \ \ \ Giorgos Tolias$^2$\ \ \ \ Yannis Avrithis$^{1}$\ \ \ \ Teddy Furon$^{1}$\ \ \ \ Ond{\v r}ej Chum$^{2}$\\
{\fontsize{11}{13}\selectfont$^1$Inria Rennes\ \ \ \ \ \ $^2$VRG, FEE, CTU in Prague}\\
{\fontsize{10}{11}\selectfont \texttt{\{ahmet.iscen,ioannis.avrithis,teddy.furon\}@inria.fr}} \\
{\fontsize{10}{11}\selectfont \texttt{\{giorgos.tolias,chum\}@cmp.felk.cvut.cz}} 
}

\maketitle
\newcommand{\head}[1]{{\smallskip\noindent\bf #1}}
\newcommand{\mypar}[1]{\noindent \textbf{#1}}
\newcommand{\ip}[2]{{#1}^{\top}{#2}}

\def \P    {\mathbf P}
\def \E    {\mathbb E}
\def \Var    {\mathbb V}
\def \real {\mathbb R}
\def \Xset {\mathcal X}
\def \Yset {\mathcal Y}
\def \Rset {\mathcal R}
\def \Qset {\mathcal Q}
\def \x{\mathbf{x}}
\def \g{\mathbf{g}}
\def \w{\mathbf{w}}
\def \q{\mathbf{q}}
\def \b{\mathbf{b}}
\def \r{\mathbf{r}}
\def \m{\mathbf{m}}
\def \h{\mathbf{h}}
\def \y{\mathbf{y}}
\def \f{\mathbf{f}}
\def \z{\mathbf{z}}
\def \Y{\mathbf{Y}}
\def \A{\mathbf{A}}
\def \B{\mathbf{B}}
\def \C{\mathbf{C}}
\def \D{\mathbf{D}}
\def \G{\mathbf{G}}
\def \Q{\mathbf{Q}}
\def \H{\mathbf{H}}
\def \X{\mathbf{X}}
\def \S{\mathbf{S}}
\def \Z{\mathbf{Z}}
\def \V{\mathbf{V}}
\def \W{\mathbf{W}}
\def \N{\mathbf{N}}
\def \M{\mathbf{M}}
\def \K{\mathbf{K}}
\def \k{\mathbf{k}}

\def \cL {\mathcal L}

\def \one{\mathbf{1}}
\def \zero{\mathbf{0}}
\def \diag{\operatorname{diag}}

\def \sim{s}

\def \PX{\mathbf{P}_{\X}}
\def \PZ{\mathbf{P}_{\Z}}
\def \PZt{\mathbf{P}_{\Z^{\bot}}}
\def \PZW{\boldsymbol{\Pi}_{\Z,\W}}
\def \Real{\mathbb{R}}
\def \U{\mathbf{U}}
\def \u{\mathbf{u}}
\def \v{\mathbf{v}}
\def \s{\mathbf{s}}
\def \c{\mathbf{c}}
\def \1{\mathbf{1}}
\def \I{\mathbf{I}}
\def \a{\mathbf{a}}
\def \mub{\boldsymbol{\mu}}
\def \Ab{\bar{\mathbf{A}}}
\def \Bb{\bar{\mathbf{B}}}
\def \Cb{\bar{\mathbf{C}}}
\def \Db{\bar{\mathbf{D}}}
\def \alb{\boldsymbol{\alpha}}
\def \Pfn {\P_\text{fn}}
\def \Pfp {\P_\text{fp}}

\newcommand{\fix}{\marginpar{FIX}}
\newcommand{\new}{\marginpar{NEW}}

\def\sssp{\hspace{1pt}}
\def\ssp{\hspace{3pt}}
\def\msp{\hspace{5pt}}
\def\bsp{\hspace{12pt}}

\def \nbitsketch {P}
\def \nbitscalar {D}
\def \MaxIter {50}
\def \pinvc {\mathsf{pinv}}
\def \sumc {\mathsf{sum}}
\newcommand{\norm}[1]{\left\lVert#1\right\rVert}

\def\etal{\textrm{et al}.\,}
\def\ie{\emph{i.e.}~}
\def\vs{\emph{vs.}~}

\def\defeq{\mathrel{\mathop:}=}

\newcommand*\rfrac[2]{{}^{#1}\!/_{#2}}

\newcommand{\alert}[1]{{\color{red}{#1}}}
\newcommand{\ahmet}[1]{\textcolor{blue}{#1}}
\newcommand{\giorgos}[1]{\textcolor{green}{#1}}

\newcommand{\os}[1]{\textbf{#1}}
\newcommand{\ns}[1]{\textbf{\textcolor{red}{#1}}}

\newcommand{\T}{{\!\top}}
\newcommand{\mT}{{-\!\top}}

\begin{abstract}
Query expansion is a popular method to improve the quality of image retrieval with both conventional and CNN representations. It has been so far limited to global image similarity.
This work focuses on diffusion, a mechanism that captures the image manifold in the feature space.
The diffusion is carried out on descriptors of overlapping image regions rather than on a global image descriptor like in previous approaches.
An efficient off-line stage allows optional reduction in the number of stored regions. In the on-line stage, the proposed handling of unseen queries in the indexing stage removes additional computation to adjust the precomputed data.
We perform diffusion
through a sparse linear system solver, yielding
practical query times well below one second.

Experimentally,
we observe
a significant boost in performance of image retrieval with compact CNN descriptors
on standard benchmarks, especially when the query object covers only a small part of the image. Small objects
have been
a common failure case of CNN-based retrieval.

\comment{
Deep learning has been recently shown successful in extracting powerful features for image retrieval. However, it has not shown yet its full potential, especially in capturing small details. One the other hand, query expansion has been limited to global image image similarity so far. We argue that CNN representations offer the opportunity to capture local information in this context with only a limited additional amount of space, compared to conventional ones.

In this paper, an image is described by a small set of deep learning features extracted from image regions. We introduce a \emph{regional diffusion mechanism} for query expansion and re-ranking. It captures up to one distinct manifold in the feature space per query region at the same cost as global diffusion. Contrary to previous work, we elegantly handle new queries not in the dataset and efficiently use a sparse linear system solver.
This brings practical query times well below one second. It provides impressive performance gains compared to global diffusion, conventional query expansion, and state of the art baselines. This is especially true when looking for small objects.

}

\end{abstract}

\vspace{-2ex}
\section{Introduction}
\label{sec:intro}

\begin{figure}[t]
\vspace{-2ex}
\centering
\includegraphics[width=0.35\textwidth]{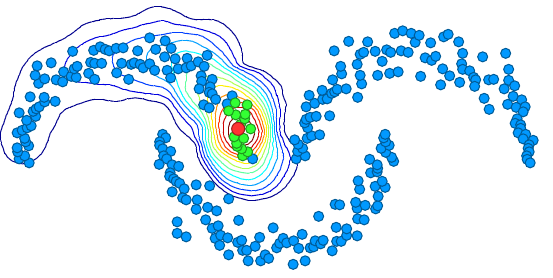} \\
(a) single query \\
\includegraphics[width=0.35\textwidth]{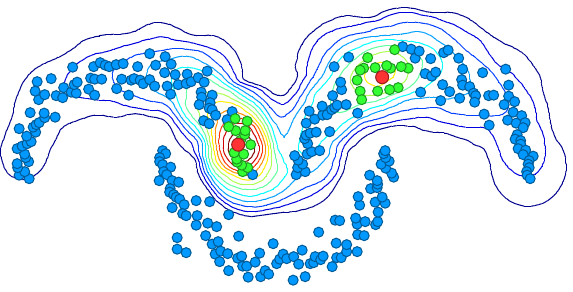} \\
(b) multiple queries
\caption{Diffusion on a synthetic dataset in $\real^2$. Dataset points, query points and their $k$-nearest neighbors are shown in blue, red, and green respectively.
Contour lines correspond to ranking scores after diffusion.
In this work, points are region descriptors.
\label{fig:bananas}}
\vspace{-10pt}
\end{figure}

Object search is a key tool behind a number of applications like content based image collection browsing~\cite{WL11,MCM13}, visual localization~\cite{SWLK12,AGT+15}, and 3D reconstruction~\cite{HSDF15,SRCF15}.
Many applications benefit from retrieving images taken from various viewing angles and under different illumination, \eg more information for the user while browsing, localization in day and night, and complete 3D models. Each image is represented
by one or more descriptors
designed or learned to exhibit a certain degree of invariance to imaging conditions. Retrieval is formulated as a nearest neighbor search in the descriptor space, performed by approximate methods%
~\cite{ML14,JDS11,KA14,BL16}.

While collections of local descriptors provide good invariance, global descriptors like VLAD~\cite{JDSP10} have smaller memory footprint, but are more prone to locking onto the clutter.
This mainly holds when the queried object covers only a small part of the image.
In case of global CNN descriptors, the invariance is partially designed by global max~\cite{ARSM+14,TSJ15} or sum~\cite{KMO15,BL15} pooling layers or multi-scale querying~\cite{GARL16b},
and partially learned by the choice of the training data.
Robustness to background clutter is improved by computing descriptors over object proposals~\cite{MoBa15,GARL16,XHZ+15} or over a fixed grid of regions~\cite{TSJ15}. Better performance is observed at a cost of increased memory footprint~\cite{RSAC14}.

In image collections, objects
are depicted  in various conditions.
As a consequence, query and relevant images are often connected by a sequence
of images, where consecutive images are similar.
The descriptors of these images form a manifold in the descriptor space.
Even though the images of the sequence contain the same object, 
the descriptors may be completely unrelated after a certain point.

This idea has been first exploited by Chum \etal~\cite{CPSIZ07} who introduce query expansion.
The average query expansion (AQE) is now used as a standard tool in image retrieval, due to its efficiency and significant performance boost. However, AQE only explores the neighborhood of very similar images.
Recursive and scale-band recursive methods~\cite{CPSIZ07} further improve the results by explicitly crawling the image manifold. This is at a cost of increased query time.

Query expansion exploits the manifold of images at query time---starting from nearest neighbors of the query and using these neighbors to issue new queries.
On the other hand, {\em diffusion}~\cite{PBM+99,ZWG+03,DB13} is based on a neighborhood graph of the dataset that is constructed off-line and efficiently uses this information at query time to search on the manifold in a principled way.

We make the following contributions:
\vspace{-4pt}
\begin{itemize}
\setlength\itemsep{-2pt}
\item
We introduce a \emph{regional diffusion mechanism}, which handles one or more query vectors at the same cost.  There is one vector per region and a few regions per image so that constructing and storing the graph is tractable. This approach significantly improves retrieval of small objects and cluttered scenes.

\item In diffusion mechanisms~\cite{PBM+99,ZWG+03,DB13}, query vectors are usually part of the dataset and available at the indexing stage. A novel approach to \emph{unseen queries} with no computational overhead is proposed.

\item Though a closed form solution is known to exist, it has been explicitly avoided so far~\cite{DB13}. We show that the commonly used alternative is in fact a well known iterative linear system solver.
Since the relevant matrix is sparse and positive definite, the \emph{conjugate gradient} method is more efficient resulting in practical query times well below one second.

\item To study the dependence of performance on relative object size, we experiment on INSTRE dataset~\cite{WJ15}, which has not received much attention so far. We propose a new evaluation protocol that is in line with other well known datasets and provide a rich set of baselines to facilitate future comparisons.

\end{itemize}

Searching in parallel in more than one manifolds via diffusion and using the nearest neighbors of unseen queries are illustrated in Figure~\ref{fig:bananas}.

The remaining text is structured as follows. Sections~\ref{sec:related} and~\ref{sec:diffusion} discuss related work and background respectively, focusing on diffusion mechanisms.
Sections~\ref{sec:method} and~\ref{sec:experiments} present our contributions in detail and the experimental body.
\section{Related work}
\label{sec:related}
This section discusses existing query expansion or re-ranking methods. We also review the concept of diffusion in computer vision and image retrieval in particular. Apart from AQE~\cite{CPSIZ07}, none of these methods has been applied to retrieval in the context of convolutional features.

\textbf{Query expansion.}
A variety of methods~\cite{CPSIZ07,CMPM11,TJ14} employ local features and are well adapted to the Bag-of-Words model~\cite{SZ03}. Others are generic and applicable on any global image representation~\cite{JHS07,DGBQG11,AZ12,SLBW14,DJAH14}. In both cases, ranking is performed on the image level. Extension to regional level is not always straightforward. If even possible, such an extension would come at a significant cost, as each query region would need to be treated independently. This is unlike our regional diffusion mechanism, which has a fixed cost with respect to the number of query regions.

\textbf{Diffusion.}
We are focusing on diffusion mechanisms, which propagate similarities through a pairwise affinity matrix~\cite{DB13,PBM+99}. They are applied to many computer vision problems, such as semi-supervised classification~\cite{ZBL+03}, seeded image segmentation~\cite{Grad06}, saliency detection~\cite{LMV14,CZHZ16}, clustering~\cite{Donoser13} and image retrieval~\cite{JBa08,EKG10,YKL09,DB13,XTZ+14}.

The power of such methods lies in capturing the intrinsic manifold structure of the data~\cite{ZBL+03}. The popular PageRank algorithm~\cite{PBM+99} was originally used to estimate the importance of web pages by exploiting their links in a graph structure. Our retrieval scenario comes closer to its so called personalized~\cite{PBM+99} or query dependent versions~\cite{RD01}, where the final ranking both respects the data manifold and the similarity to a number of query vectors.

Diffusion is used for retrieval of general scenes or shapes of particular objects~\cite{JBa08,EKG10,YKL09,DB13}. It can also fuse multiple feature modalities~\cite{ZYCYM12,YMD15} by jointly modeling them on the same graph. In these approaches, images are the nodes of the graph with edges established given a pairwise similarity measure. We differentiate by defining a graph of image \emph{regions} linked based on region similarities while performing a single pseudo random walk for multiple query regions. Diffusion with regional similarity has been investigated before, but only to define image level affinity~\cite{ZNC16}, to aggregate local features~\cite{FO15}, or to handle bursts~\cite{GXZPT16}.

Donoser and Bischof~\cite{DB13} review a number of diffusion mechanisms for retrieval. They focus on iterative solutions arguing that closed form solutions, when existing, are impractical due to inversion of large matrices.
We rather focus on a closed form solution computed approximatively with an iterative method that is particularly designed for this problem and show that this approach is faster.
\section{Ranking with diffusion}
\label{sec:diffusion}
Diffusion in the work of Donoser and Bischof~\cite{DB13} denotes a mechanism spreading the query similarities over the manifolds composing the dataset. This is only weakly related to continuous time diffusion process or random walks on graph. We mainly follow Zhou \etal~\cite{ZWG+03} below.

\head{Affinity matrix.}
Given a dataset $\Xset := \{\x_1, \dots, \x_n\} \subset \real^d$, we define the \emph{affinity matrix} $A = (a_{ij}) \in \real^{n\times n}$ having as elements the pairwise similarities between points of $\Xset$:
\vspace{-4pt}
\begin{equation}
	a_{ij} := \sim(\x_i, \x_j), \quad\forall (i,j) \in [n]^2,
\label{equ:similarity_pairwise}
\vspace{-4pt}
\end{equation}
where $[n]:=\{1, \dots, n \}$ and $\sim: \real^d \times \real^d \rightarrow \real$ is a similarity measure assumed to be symmetric ($A = A^\T$), positive ($A > 0$), and with zero self-similarities ($\diag(A) = \zero$).

Matrix $A$ is the adjacency matrix of a weighted undirected graph $G$ with vertices $\Xset$. The degree matrix of the graph is $D := \diag(A\one_n)$, \ie a diagonal matrix with the row-wise sum of $A$. The Laplacian of the graph is defined as
$L := D - A$. It is usual to symmetrically normalize these matrices, for instance,
\vspace{-4pt}
\begin{equation}
	S := D^{-1/2} A D^{-1/2},
\label{equ:transition}
\vspace{-4pt}
\end{equation}
for the affinity matrix and $\cL := I_n - S$ for the Laplacian, where $I_n$ denotes the identity matrix of size $n$. Matrices $L, \cL$ are positive-semidefinite~\cite{Chun97}.

\head{Diffusion.}
In the work of Zhou \etal~\cite{ZWG+03}, a vector $\y = (y_i) \in \real^n$ specifies a set of query points in $\Xset$, with $y_i = 1$ if $\x_i$ is a query and $y_i = 0$ potherwise.
The objective is to obtain a ranking score $f_i$ for each point $\x_i \in \Xset$, represented as vector $\f = (f_i) \in \real^n$.

We focus on a particular diffusion mechanism that, given an initial vector $\f^0$, iterates according to
\vspace{-4pt}
\begin{equation}
	\f^t = \alpha S \f^{t-1} + (1 - \alpha) \y.
\label{equ:diffusion_iter}
\vspace{-4pt}
\end{equation}
If $S$ is a transition matrix and $\y$ a $\ell^1$ unit vector, this defines the following `random walk' on the graph: with probability $\alpha$ one jumps to an adjacent vertex according to distribution stated in $S$, and with $1-\alpha$ to a query point uniformly at random. In this fashion, points spread their ranking score to their neighbors in the graph.
The benefit 
is the ability to capture the intrinsic manifold structure represented by the affinity matrix and to combine multiple query points.

Assuming $0 < \alpha < 1$, Zhou \etal~\cite{ZBL+03,ZWG+03} show that sequence $\{\f^t\}$ defined by~(\ref{equ:diffusion_iter}) converges to
\vspace{-4pt}
\begin{equation}
    \f^\star = (1-\alpha) \cL_\alpha^{-1} \y
\label{equ:diffusion_closed}
\vspace{-4pt}
\end{equation}
where $\cL_\alpha := I_n-\alpha S$ is  positive-definite.
This follows since $\cL_\alpha = \alpha \cL + (1 - \alpha) I_n \succ \alpha \cL \succeq 0$. In this work, we focus on the \emph{closed form} solution~(\ref{equ:diffusion_closed}) rather than its intuitive derivation from iterative process~(\ref{equ:diffusion_iter}).

\head{Relation to other approaches.}
A diffusion mechanism also appears in seeded image segmentation~\cite{Grad06}, where query points correspond to labeled pixels (seeds) and database points to the remaining unlabeled pixels. This problem is equivalent to semi-supervised classification~\cite{ZBL+03}.
In our context, the approach of Grady~\cite{Grad06} decomposes $\f=(\f_d^\T,\f_q^\T)^\T$ for the scores of the query (fixed $\f_q$) and database (unknown $\f_d$) points. Diffusion interpolates $\f_d$ from $\f_q$ by minimizing, w.r.t. $\f_d$, the quadratic cost $\sum_{i,j}a_{ij}(f_i-f_j)^2 = \f^\T L\f$ to enforce that neighboring points should have similar scores. By decomposing $L = [L_d,\,-S_{qd};-S_{qd}^\T,\,L_q]$,
it is shown~\cite{Grad06} that the solution fulfills $L_d\f_d =\y$ with $\y=S_{qd}^\T\f_q$.
In our setup, $L_d$ would be singular, preventing us to single out a solution $\f_d^{\star}$.
Yet, it is easy to show that the minimizer of the cost $\alpha \f^\T L \f +(1-\alpha)\|\f\|^2$ has a similar expression to~(\ref{equ:diffusion_closed}). The regularization term singles out a solution by forcing $\f$ to be zero in subgraphs not connected to any query point. The details are omitted for brevity.

\head{Local constraints.}
Donoser and Bischof have extensively investigated various constructions of affinity matrices in the context of image retrieval~\cite{DB13}. Our work uses matrix~\eqref{equ:transition}, which is found to be the most effective in their work, and is also used by Zhou \etal~\cite{ZBL+03}. Further, to handle noise and outliers, we adopt a locally constrained random walk~\cite{KDB09} where only pairs of points that are reciprocal (mutual) nearest neighbors are kept as edges in the graph. In particular, given $\z \in \real^d$, let
\vspace{-4pt}
\begin{equation}
	\sim_k(\x | \z) =
	\begin{cases}
		\sim(\x, \z), & \text{if } \x \in \text{NN}_k(\z) \\
		0,            & \text{otherwise}
	\end{cases}
\label{equ:knn_similarity}
\vspace{-4pt}
\end{equation}
be the similarity of $\x \in \Xset$ given $\z$, that is, restricted to the $k$ nearest neighbors $\text{NN}_k(\z)$ of $\z$ in $\Xset$. Then,
\vspace{-4pt}
\begin{equation}
	\sim_k(\x, \z) = \min \{ \sim_k(\x | \z), \sim_k(\z | \x) \}
\label{equ:knn_affinity}
\vspace{-4pt}
\end{equation}
equals $s(\x, \z)$ if $\x, \z$ are the $k$-nearest neighbors of each other in $\Xset$, and zero otherwise. We use similarity function $\sim_k$ to construct affinity matrix $A$ like in~\eqref{equ:similarity_pairwise}.

\section{Method}
\label{sec:method}

This section describes our contributions on image retrieval: handling new query points not in the dataset, searching for multiple regions with a single diffusion mechanism, and efficiently computing the solution.

\subsection{Handling new queries}
\label{sec:queryNodes}

In prior work on diffusion, a query point $\q$ is considered to be contained in the dataset $\Xset$~\cite{ZBL+03, DB13}. This does not hold in a retrieval scenario, but a query can be included in the dataset graph at query time~\cite{ZYCYM12} as follows. The $k$ nearest neighbors $\text{NN}_k(\q)$ of $\q$ in $\Xset$ are found and reciprocity is checked. The rows and columns of the affinity matrix $A$ corresponding to $\text{NN}_k(\q)$ are updated to maintain~\eqref{equ:knn_affinity} in the presence of $\q$, and $A$ is augmented by appending an extra row and column for $\q$. Matrix $S$ is computed by normalizing $A$~\eqref{equ:transition}. Finally, vector $\y$ indicates that $\q$ is a query. Generalizing to multiple query points is straightforward.

Even if we ignore the time needed for the above computation, we argue that locking, modifying and augmenting the entire affinity matrix for each query is not acceptable in terms of space requirements\footnote{Imagine the case of multiple users querying at the same time; a different matrix per query is required. Also, updating mutual neighbors requires $k$-NN lists which are not available any longer.}. We introduce here an alternative method which defines vector $\y$ in a new way rather than modifying $A$. Qualitatively, instead of searching for $\q$, we are searching for its neighbors $\text{NN}_k(\q)$, appropriately weighted. In particular, we define $\y$ as
\vspace{-4pt}
\begin{equation}
	y_i = \sim_k(\x_i | \q), \quad \forall i \in [n].
\label{equ:single_query_vector}
\vspace{-4pt}
\end{equation}
Our motivation for this choice is detailed in Section~\ref{sec:regDiff}  including the more general case of multiple query points. Figure~\ref{fig:bananas} shows a toy 2-dimensional example of diffusion, where the $k$-nearest neighbors to each query point taken into account in~\eqref{equ:single_query_vector} are depicted.
It is evident that multiple manifolds are captured when multiple queries are issued.
Section~\ref{sec:experiments} experimentally shows improved performance compared to the conventional approach.

\subsection{Regional diffusion}
\label{sec:regDiff}
The diffusion mechanism described so far is applicable to image retrieval when database and query images are globally represented by single vectors.
We call this \emph{global diffusion} in the rest of the paper. Unlike the traditional representation with local descriptors~\cite{SZ03,PCISZ07}, global diffusion fits perfectly with the early CNN-based global features~\cite{BL15,KMO15,RTC16}.

Global features still fail under severe occlusion or when the object of interest is small. Local CNN features from multiple image regions have been investigated for this purpose, either aggregated~\cite{GWGL14,TSJ15} or represented as a set~\cite{RSAC14}. Given a query image, the latter means that one searches for each query feature individually.

Fortunately, diffusion as defined in section~\ref{sec:diffusion} can already handle multiple queries. In the following, an image is represented by a set $X_i \subset \real^d$ of $m$ points, one for each region. Dataset $\Xset$ is the union of such sets over all images; $n$ still denotes its size. The query image is also represented by a set $Q$ of $m$ points.
Each region feature is a point possibly lying on a different manifold. We discuss below the new definition of vector $\y$ and the combination of individual region ranking scores into a single score per image. We call this mechanism \emph{regional diffusion}.

\head{Specifying queries.}
In the conventional approach where query points are in the dataset,
one directly applies~\eqref{equ:diffusion_iter} with $\y\in\{0,1\}^{n+m}$ with $m$ non-zero elements indicating the query points. This situation resembles the personalized PageRank~\cite{PBM+99}. However, it is simpler to keep $A$ as an $n\times n$ affinity matrix and to set $\y\in\real^n$ as
\vspace{-4pt}
\begin{equation}
	y_i:= \sum_{\q \in Q} \sim_k(\x_i | \q), \quad\forall i \in [n].
\label{equ:multi_query_vector}
\vspace{-4pt}
\end{equation}
Each dataset point $\x_i$ is assigned a scalar that is the sum of similarities over all query points $\q$ for which $\x_i$ appears in the corresponding $k$-nearest neighbor set $\text{NN}_k(\q)$, and zero if it appears in no such set.

\head{Derivation.}
Our work is inspired by the analysis in the work of Grady~\cite{Grad06} that we apply to the diffusion mechanism of Zhou \etal~\cite{ZWG+03}, where query points $Q$ are in the dataset. We decompose the quantities in~(\ref{equ:diffusion_iter}) as $\f = (\f_d^\T, \f_q^\T)^\T$, with $\f_d\in\real^n$ and $\f_q\in\real^m$,
\vspace{-2pt}
\begin{equation}
	S = \left( \begin{array}{cc} S_d & B_{dq} \\ B_{qd} & S_q \end{array} \right),
\label{equ:aff_decomp}
\vspace{-2pt}
\end{equation}
and $\y = (\zero_n^\T, \one_m^\T)^\T$. Subscripts $d,q$ denote data and query respectively. Then,~(\ref{equ:diffusion_iter}) is written as
\begin{align}
	\f^t_d & = \alpha S_d \f^{t-1}_d + \alpha B_{dq} \f^{t-1}_q
	\label{equ:diff_sys_data} \\
	\f^t_q & = \alpha B_{qd} \f^{t-1}_d + \alpha S_q \f^{t-1}_q + (1-\alpha) \one_m.
	\label{equ:diff_sys_query}
\end{align}
Provided this system converges, the data part satisfies
\begin{equation}
\f_d^\star \propto \cL_\alpha^{-1}B_{dq}\one_m
\label{eq:OurDiffusion}
\end{equation}
if $\f^\star_q \propto \one_m$, $S_q=\zero_{m\times m}$ and $B_{qd}=\zero_{m\times n}$. In words, the query points are perfectly retrieved, they are dissimilar to each other, and the graph is indeed directed with query regions pointing to dataset regions, but the reverse is not allowed. 
Comparing~\eqref{eq:OurDiffusion} with~\eqref{equ:diffusion_closed}, it follows that $B_{dq} \one_m$ is a good choice for $\y$. Since $B_{dq}$ stores the similarities between the dataset and the query points, this analysis justifies the single query~\eqref{equ:single_query_vector} and the multiple queries~\eqref{equ:multi_query_vector} cases.

\head{Diffusion.}
Given this definition of $\y$, diffusion is now performed on dataset $\Xset$,
jointly for all query points in $Q$. Affinities of multiple query points are propagated in the graph in a single process at no additional cost compared to the case of a single query point. Here we are excluding the additional cost of computing $\y$ itself in~(\ref{equ:multi_query_vector}) compared to~(\ref{equ:single_query_vector}).
This search takes place in all related work.
We also do not discuss how to make this search more efficient in space and time~\cite{BL16}, which is beyond the scope of this work.

Figure~\ref{fig:bananas} illustrates the diffusion on single and multiple query points. The contour lines show the ranking score any point on the plane would be assigned given the query point(s). It is evident that multiple manifolds are captured when multiple queries are issued.

\head{Pooling.}
After diffusion, each image is associated with several elements of the ranking score vector $\f^\star$, one for each point $\x$ in $X \subset \Xset$. A simple way to combine these scores is to define the score of image $X$ as
\vspace{-4pt}
\begin{equation}
	f(X) = \sum_{j \in [m]} w_j f^\star_{i_X(j)},
\label{equ:gm_pool}
\vspace{-4pt}
\end{equation}
where $i_X(j)$ is the index of the $j$-th point of $X$ in the dataset $\Xset$ and $\w=(w_j)$ a weighting vector. The latter is defined as $\w=\one_m$ for \emph{sum pooling} and, assuming $m<d$,
\vspace{-4pt}
\begin{equation}
	\w = (\Phi \Phi^\T + \lambda I_m)^{-1} \one_m
\label{equ:gm_weights}
\vspace{-4pt}
\end{equation}
for \emph{generalized max pooling} (GMP)~\cite{MP14,IFGRJ14}, where $\Phi=(\x_{i_X(1)}^\T,\ldots,\x_{i_X(m)}^\T)^\T$ and $\lambda\in\real^+$ is a regularization parameter.
Our experiments show that GMP always outperforms sum pooling.

\subsection{Efficient solution}
\label{sec:appProp}

Iteration~(\ref{equ:diffusion_iter}) works well in practice but is slow at large scale. Taking the closed-form solution~(\ref{equ:diffusion_closed}) literally, one may be tempted to compute the inverse $\cL_\alpha^{-1}$ offline, but this matrix is not sparse like $\cL_\alpha$. We propose a more efficient solution by making the connection to linear system solvers.

\head{Diffusion is an iterative solver.}
Eq.~\eqref{equ:diffusion_iter} can be seen as an iteration of the \emph{Jacobi} solver~\cite{Hack94}.
Given a linear system $A\x = \b$\footnote{We adopt the standard linear system notation in this section; matrix $A$ is not to be confused with our affinity matrix defined in~(\ref{equ:similarity_pairwise}).}, Jacobi decomposes $A$ as $A = \Delta + R$ where $\Delta = \diag(A)$. 
It then iterates according to
\vspace{-4pt}
\begin{equation}
	\x^t = \Delta^{-1} (\b - R \x^{t-1}).
\label{eq:jacobi}
\vspace{-4pt}
\end{equation}
In our case, $\x = \f$, $\b = (1-\alpha) \y$, and $A = \cL_\alpha = I - \alpha S$. It follows that $\Delta = I_n$ and $R = -\alpha S$, so that
\vspace{-4pt}
\begin{equation}
	\f^t = \alpha S \f^{t-1} + (1-\alpha) \y.
\label{eq:diffusion_item_der}
\vspace{-4pt}
\end{equation}
We have just re-derived~(\ref{equ:diffusion_iter}). Note that a sufficient condition for Jacobi's convergence is that matrix $A$ is strictly diagonally dominant, \ie $|a_{ii}| > \sum_{j \ne i} a_{ij}$ for $i \in [n]$. It is easily checked that $\cL_\alpha$ does satisfy this condition by construction, given that $0 < \alpha < 1$. This provides an alternative proof of the main result of Zhou \etal~\cite{ZBL+03}.

\head{Conjugate gradient} (CG)~\cite{NoWr06} is the method of choice for solving linear systems like ours
\vspace{-4pt}
\begin{equation}
	\cL_\alpha \f = (1-\alpha) \y,
\label{eq:diffusion_system}
\vspace{-4pt}
\end{equation}
where $\cL_\alpha$ is
positive-definite, and in particular for graph-related problems~\cite{Vish12}. It has been used for random walk problems~\cite{Grad06},
but not diffusion-based retrieval according to our knowledge. In fact, the linear system formulation
has been explicitly avoided in this context~\cite{DB13}.

Here we argue, as in~\cite{LaMe04}, that it is the solution of~(\ref{eq:diffusion_system}) that we seek, rather than the path followed by iteration~(\ref{equ:diffusion_iter}).
However, we use CG to approximate this solution, since matrix $\cL_\alpha$ is indeed positive-definite. At each iteration, CG minimizes the quadratic function $\phi(\x) = \frac{1}{2} \x^\T A \x - \x^\T \b$ in a particular direction by analytically computing the optimal step length. More importantly, the direction chosen at each iteration is conjugate to previous ones. Thus, any update of $\x$ along this direction does not destroy the optimality reached in the entire subspace considered thus far.

Contrary to other iterative methods including~\eqref{eq:diffusion_item_der}, CG is guaranteed to terminate in $n$ steps. Remarkably, it provides good approximations in very few steps.

\head{Normalization is preconditioning.} Finally, a standard improvement is preconditioning, \ie, solving a related system with $A$ replaced by $C^{-1} A C^\mT$, a matrix satisfying a weak condition like its eigenvalues being clustered. Unfortunately, finding an appropriate matrix $C$ can be quite complex~\cite{Vish12}. We observe that  normalization~\eqref{equ:transition} \emph{is} preconditioning.%
Indeed, we could equally consider matrix $L_\alpha = D - \alpha A = \alpha L + (1 - \alpha) I \succ 0$ and solve the linear system
\begin{equation}
	L_\alpha (D^{-1/2} \f) = (1 - \alpha) (D^{1/2} \y)
\label{eq:diffusion_system_pre}
\end{equation}
instead, which is equivalent to~(\ref{eq:diffusion_system}). By normalizing $L_\alpha$ into $\cL_\alpha$, we are actually performing preconditioning with $C = \diag(L_\alpha)^{1/2}$. This is a simple form of symmetric preconditioning, known as \emph{diagonal scaling} or \emph{Jacobi}~\cite{Tref97}. It improves convergence, be it for CG or diffusion~\eqref{equ:diffusion_iter}.

\subsection{Scaling up}
\label{sec:largescale}
Despite the efficient solution described in the previous section, there are still issues concerning space and offline pre-processing at large scale. We address these issues here.

\head{Compact representation.}
At large scale, the number of region features per database image should be kept as low as possible.
For this reason, we learn a Gaussian Mixture Model (GMM) on the original features of each database image and represent the image by the unit normalized means. 
This is an even more natural choice when dealing with overlapping regions (see Section~\ref{sec:experiments}). 
As a result, it decreases the number of region features and their redundancy.

\head{The off-line construction}
of the affinity matrix
is quadratic in the number of vectors in the database and might not be tractable at large scale.
We employ the efficient and approximate $k$-NN graph construction method by Dong~\etal~\cite{DCL11}. Section~\ref{sec:experiments} shows that it is orders of magnitude faster than exhaustive search and has almost no effect on performance.

\head{Truncating the affinity matrix.}
Instead of ranking the full dataset, diffusion re-ranks an initial search. 
This baseline in our experiments is done with global descriptors and kNN search. Then we apply diffusion only on the top ranked images. We truncate the affinity matrix keeping only the rows and columns related to the regions of the top ranked images and re-normalize it according to~(\ref{equ:transition}). The cost of this step is not significant compared to the actual diffusion.

\section{Experiments}
\label{sec:experiments}

This section presents the experimental setup and investigates the accuracy of our methods for
image retrieval compared with the state-of-the-art approaches.

\subsection{Experimental Setup}
\label{sec:expSetup}
\head{Datasets.}
We use three datasets.
Two are well-known image retrieval benchmarks: Oxford Buildings~\cite{PCISZ07} and Paris~\cite{PCISZ08}.
We refer to them as Oxford5k and Paris6k.
We experiment at large-scale by adding 100k distractor images from Flickr~\cite{PCISZ07}, forming Oxford105k and Paris106k datasets.
The third corpus is the recently introduced instance search dataset called INSTRE~\cite{WJ15}.
It contains various everyday 3D or planar objects from buildings to logos with many variations such as different scales, rotations, and occlusions. Some objects cover a small part of the image, making it a challenging dataset.
It consists of 28,543 images from 250 different object classes. In particular, 100 classes with images retrieved from on-line sources, 100 classes with images taken by the dataset creators, and 50 classes consisting of pairs from the second category.
We differentiate from the original protocol~\cite{WJ15}, which uses all database images as queries.
We randomly split the dataset into 1250 queries, 5 per class, and 27293 database images, while a bounding box defines the query region\footnote{\url{http://people.rennes.inria.fr/Ahmet.Iscen/diffusion.html}}.
The query and the database sets have no overlap.
We use mean average precision (mAP) as a performance measure in all datasets.

\begin{table}[t]
\begin{center}
\footnotesize
    \begin{tabular}{|@{\ssp}l@{\ssp}|@{\ssp}c@{\ssp}|@{\ssp}c@{\ssp}|@{\ssp}c@{\ssp}|@{\ssp}c@{\ssp}|@{\ssp}c|}
    \hline
      	Pooling				& INSTRE  				& Oxf5k				& Oxf105k	    	& Par6k	    		& Par106k	  	\\ \hline \hline
      	sum				      	& 79.1  					& 92.2				&	90.6				& 96.1				& 94.4			\\
      	GMP	 		 	  			& 80.0 					& 93.2				&	91.6				& 96.5				& 94.6			\\ \hline

\end{tabular}
\vspace{2pt}
\caption{Retrieval performance (mAP) of regional diffusion with sum and generalized max pooling (GMP), with $\lambda = 1$ in~\eqref{equ:gm_weights}.
\label{tab:gmPool}}
\end{center}
\end{table}

\head{Representation.}
We employ a CNN that is fine-tuned for image retrieval~\cite{RTC16} to extract global and regional representation. In particular, this fine-tuned VGG produces 512 dimensional descriptors.
We extract regions at 3 different scales as in R-MAC~\cite{TSJ15}, while we additionally include the full image as a region.
In this fashion, each image has on average 21 regions.
The regional descriptors are aggregated and re-normalized to unit norm in order to construct the  global descriptors, which is exactly as in R-MAC.
We apply supervised whitening~\cite{RTC16} to both global and regional descriptors.
We use this network to perform all our initial experiments.
In Section~\ref{sec:comparison}, we also report scores with higher dimensional descriptors derived from the fine-tuned ResNet101~\cite{GARL16b} using the same fixed grid.

\head{Implementation details.}
We define the affinity function using a monomial kernel~\cite{TAJ13} as $\sim(\x, \z) =  \max(\x^\T\z,0)^{3}$.
The diffusion parameter $\alpha$ is always 0.99, as in the work of Zhou \etal~\cite{ZBL+03}.
The $k$-NN search required by~\eqref{equ:multi_query_vector} is assumed to access all database vectors exhaustively. Our work does not investigate how approximate search methods~\cite{ML14,JDS11,KA14,BL16,IRF16} could improve time and space consumed by this process.
After computing~\eqref{equ:multi_query_vector}, we only keep the largest $k$ values of $\y$ and set the rest to zero.

\subsection{Impact of different components}
\label{sec:retrPerf}
\begin{figure}[t]
\centering
\input{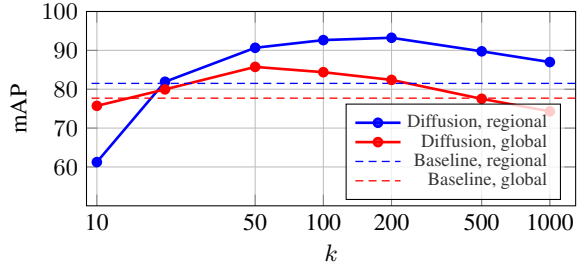}
\small
\begin{tabular}{c}
\extfig{diffKaOxf}{
\begin{tikzpicture}
\begin{axis}[%
	width=0.97\linewidth,
	height=0.5\linewidth,
	xlabel={$k$},
	ylabel={mAP},
	legend pos=south east,
   legend style={cells={anchor=east}, font =\scriptsize, fill opacity=0.8, row sep=-2.8pt},
    ymax = 100,
    ymin = 50,
    xmin = 9,
    xmax = 1300,
    grid=both,
    xmode = log,
    xtick={10,50,100,200,500,1000},
    xticklabels={10,50,100,200,500,1000},
    ytick={60,70,...,100},
 	 y label style={at={(axis description cs:0.05,.5)}},
 	 x label style={at={(axis description cs:.5,.05)}}
]
	\addplot[color=blue,     solid, mark=*,  mark size=1.5, line width=1.0] table[x=kavar, y expr={100*\thisrow{oxf5klocal}}]  \diffKa;\leg{Diffusion, regional};
	\addplot[color=red,     solid, mark=*,  mark size=1.5, line width=1.0] table[x=kavar, y expr={100*\thisrow{oxf5kglobal}}]  \diffKa;\leg{Diffusion, global};
	\addplot[color=blue, densely dashed, line width=0.5] coordinates {(0.01,81.5) (1300,81.5)}; \leg{Baseline, regional};
	\addplot[color=red, densely dashed, line width=0.5] coordinates {(0.01,77.7) (1300,77.7)}; \leg{Baseline, global};
\end{axis}
\end{tikzpicture}
}
\end{tabular}
\caption{Impact of the number of nearest neighbors $k$ in the affinity matrix.
mAP performance for global and regional diffusion on Oxford5k; baselines are R-MAC and R-match respectively.
\label{fig:diffKa}
}
\end{figure}

\head{Neighbors.}
We vary the number of nearest neighbors $k$ for constructing the affinity matrix and evaluate performance for both global and regional diffusion.
The global baseline method is $k$-NN search with R-MAC, while the regional one is the method by Razavian \etal~\cite{RSAC14}, where image regions are indexed and cross-matched.
We refer to the latter as R-match in the rest of our experiments.

Results for Oxford5k are presented in Figure~\ref{fig:diffKa}, and are consistent in other datasets.
The performance stays stable over a wide range of $k$.
The drop for low $k$ is due to very few neighbors being retrieved (where regional diffusion is more sensitive), whereas for high $k$, it is due to capturing more than the local manifold structure (where regional diffusion is superior). This behavior is consistent with the fact that small patterns appear more frequently than entire images.

We set $k=200$ for regional diffusion, and $k=50$ for global diffusion for the rest of our paper.
Since only mutual neighbors are linked, the actual number of edges per element is less:
The average number of edges per image (resp. region) is 25 (resp. 75) for global (resp. regional) diffusion, measured on INSTRE.
We set $k=200$ for the query as well in the case of the regional diffusion, while for the global one $k=10$ is needed to achieve good performance.

\head{Pooling.} We evaluate the two pooling strategies after regional diffusion in Table~\ref{tab:gmPool}.
Generalized max pooling has a small but consistent benefit in all datasets.
We use this strategy for the rest of our experiments.
Weights~\eqref{equ:gm_weights} are computed off-line and only one scalar per region is stored.

\begin{figure}[t]
\centering
\input{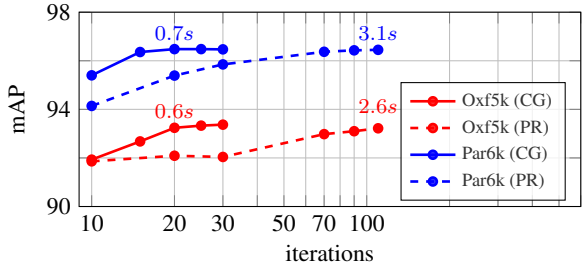}
\small
\begin{tabular}{c}
\extfig{diffTimeIt}{
\begin{tikzpicture}
\begin{axis}[%
	width=0.99\linewidth,
	height=0.5\linewidth,
	xlabel={iterations},
	ylabel={mAP},
      legend pos=south east,
      legend cell align=left,
      legend style={font=\scriptsize, fill opacity=0.8},
    ymax = 98,
    ymin = 90,
    xmin = 9,
    xmax = 600,
    grid=both,
    xmode = log,
    xtick={10,20,...,100,200,300,...,600},
    xticklabels={10,20,30,,50,,70,,,100,},
    ytick={90,92,...,100},
    yticklabels={90,,94,,98},
 	 y label style={at={(axis description cs:0.06,.5)}},
 	 x label style={at={(axis description cs:.5,.05)}},
]
	\addplot[color=red,    solid, mark=*,  mark size=1.5, line width=1.0] table[x=it, y expr={100*\thisrow{oxford5kCG}}] \diffIt;\leg{Oxf5k (CG)};
	\addplot[color=red,   dashed, mark=*, mark options={solid}, mark size=1.5, line width=1.0] table[x=it, y expr={100*\thisrow{oxford5kPR}}]  \diffIt;\leg{Oxf5k (PR)};
	\addplot[color=blue,   solid, mark=*,  mark size=1.5, line width=1.0] table[x=it, y expr={100*\thisrow{paris6kCG}}]  \diffIt;\leg{Par6k (CG)};
	\addplot[color=blue,   dashed, mark=*,  mark options={solid}, mark size=1.5, line width=1.0] table[x=it, y expr={100*\thisrow{paris6kPR}}]  \diffIt;\leg{Par6k (PR)};
	\node [above] at (axis cs:  20,  96.5) {\footnotesize \textcolor{blue}{$0.7 s$}};
	\node [above] at (axis cs:  20,  93.3) {\footnotesize\textcolor{red}{ $0.6 s$}};
	\node [above] at (axis cs:  110,  96.5) {\footnotesize \textcolor{blue}{$3.1 s$}};
	\node [above] at (axis cs:  110,  93.5) {\footnotesize\textcolor{red}{ $2.6 s$}};

\end{axis}
\end{tikzpicture}
}
\end{tabular}
\caption{mAP performance of regional diffusion \vs number of iterations for conjugate gradient (CG) and iterative diffusion~(\ref{equ:diffusion_iter}). Labels denote diffusion time.
\label{fig:diffIt}
\vspace{-4pt}
}
\end{figure}

\head{Efficient diffusion with conjugate gradient.}
We compare the iterative diffusion~(\ref{equ:diffusion_iter}) to our conjugate gradient solution.
We iterate each method until convergence.
Performance is presented in Figure~\ref{fig:diffIt} with timings measured on a machine with a 4-core Intel Xeon 2.00GHz CPU.
CG converges in as few as 20 iterations, which are also faster,
while~(\ref{equ:diffusion_iter}) reaches the same performance as CG only after 110 iterations.

The average query time on Oxford5k including all stages for global baseline, regional baseline, global diffusion and regional diffusion without truncation is 0.001s, 0.321s, 0.02s, and 0.664s, respectively.

\head{Handling new queries.}
We compare our new way of handling new queries to the conventional approach that assumes queries to be part of the dataset.
Our method achieves 80.0 mAP on INSTRE compared to 77.7 achieved by the conventional approach. We therefore not only offer space improvements but also better performance,%
mainly in the case of regional diffusion.
The main difference is that $k$ nonzero elements are kept both per query region%
~(\ref{equ:multi_query_vector}) and for the entire vector $\y$. This, due to the overlapping nature of the CNN regions, may filter out incorrect neighbors.

\subsection{Large scale diffusion}
\label{sec:largescaleExp}

\begin{figure}[t]
\vspace{-5pt}
\centering
\input{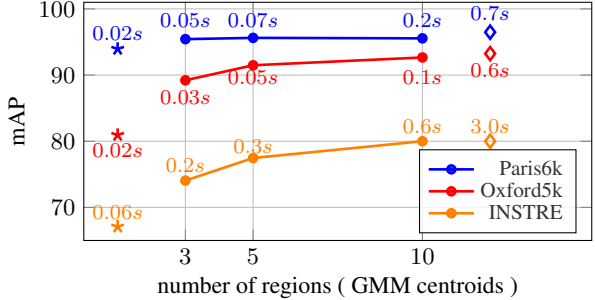}
\small
\begin{tabular}{c}
\extfig{diffGmm}{
\begin{tikzpicture}
\begin{axis}[%
	width=1\linewidth,
	height=0.57\linewidth,
	xlabel={number of regions ( GMM centroids )},
	ylabel={mAP},
	legend pos=south east,
    legend style={cells={anchor=east}, font =\footnotesize, fill opacity=0.8, row sep=-2.5pt},
    ymax = 101,
    ymin = 65,
    xmin = 0,
    xmax = 15,
    grid=both,
    xtick={3,5,10},
    ytick={60,70,...,100},
 	 y label style={at={(axis description cs:0.05,.5)}},
 	 x label style={at={(axis description cs:.5,.05)}}
]
	\addplot[color=blue,     solid, mark=*,  mark size=1.5, line width=1.0] table[x=l, y expr={100*\thisrow{paris6k}}]  \gmmL;\leg{Paris6k};
	\addplot[color=red,    solid, mark=*,  mark size=1.5, line width=1.0] table[x=l, y expr={100*\thisrow{oxford5k}}]  \gmmL;\leg{Oxford5k};
	\addplot[color=orange,     solid, mark=*,  mark size=1.5, line width=1.0] table[x=l, y expr={100*\thisrow{instre}}]  \gmmL;\leg{INSTRE};
	\addplot[color=red, mark=star, only marks, mark size = 2.5, line width = 1] coordinates {(1,80.94)};
	\addplot[color=blue, mark=star, only marks, mark size = 2.5, line width = 1] coordinates {(1,93.95)};
	\addplot[color=orange, mark=star, only marks, mark size = 2.5, line width = 1] coordinates {(1,67.08)};
	\addplot[color=red, mark=diamond, only marks, mark size = 2.5, line width = 1] coordinates {(12,93.24)};
	\addplot[color=blue, mark=diamond, only marks, mark size = 2.5, line width = 1] coordinates {(12,96.48)};
	\addplot[color=orange, mark=diamond, only marks, mark size = 2.5, line width = 1] coordinates {(12,80.0)};
	\node [above] at (axis cs:  1,  94) {\footnotesize \textcolor{blue}{$0.02 s$}};
	\node [above] at (axis cs:  3,  96) {\footnotesize\textcolor{blue}{ $0.05 s$}};
	\node [above] at (axis cs:  5,  96) {\footnotesize \textcolor{blue}{$0.07 s$}};
	\node [above] at (axis cs:  10,  96) {\footnotesize \textcolor{blue}{$0.2 s$}};
	\node [above] at (axis cs:  12,  97) {\footnotesize \textcolor{blue}{$0.7 s$}};
	\node [below] at (axis cs:  1,  81) {\footnotesize \textcolor{red}{$0.02 s$}};
	\node [below] at (axis cs:  3,  89) {\footnotesize \textcolor{red}{$0.03 s$}};
	\node [below] at (axis cs:  5,  92) {\footnotesize \textcolor{red}{$0.05 s$}};
	\node [below] at (axis cs:  10,  92) {\footnotesize \textcolor{red}{$0.1 s$}};
	\node [below] at (axis cs:  12,  93) {\footnotesize \textcolor{red}{$0.6 s$}};
	\node [above] at (axis cs:  1,  67) {\footnotesize \textcolor{orange}{$0.06 s$}};
	\node [above] at (axis cs:  3,  74) {\footnotesize \textcolor{orange}{$0.2 s$}};
	\node [above] at (axis cs:  5,  77) {\footnotesize \textcolor{orange}{$0.3 s$}};
	\node [above] at (axis cs:  10,  80) {\footnotesize \textcolor{orange}{$0.6 s$}};
	\node [above] at (axis cs:  12,  80) {\footnotesize \textcolor{orange}{$3.0 s$}};

\end{axis}
\end{tikzpicture}
}
\end{tabular}
\caption{ mAP performance for varying number of regional descriptors after learning a GMM per image. Symbol $\star$ denotes global diffusion, and $\diamond$ to the default number of regions  (21) per image. Average diffusion time in seconds is shown in text labels.
\label{fig:gmmDiff}
}
\end{figure}
\begin{figure}[t]
\centering
\input{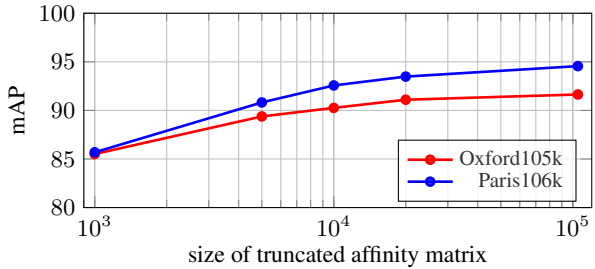}
\begin{tabular}{c}
\small
\extfig{diffTruncPtg}{
\begin{tikzpicture}
\begin{axis}[%
	width=1\linewidth,
	height=0.5\linewidth,
	xlabel={size of truncated affinity matrix},
	ylabel={mAP},
	legend pos=south east,
    legend style={cells={anchor=east}, font =\footnotesize, fill opacity=0.8, row sep=-2.5pt},
    ymax = 100,
    ymin = 80,
    xmin = 900,
    xmax = 120000,
    grid=both,
    xmode = log,
 	 y label style={at={(axis description cs:0.05,.5)}},
 	 x label style={at={(axis description cs:.5,.05)}}
]
	\addplot[color=red,     solid, mark=*,  mark size=1.5, line width=1.0] table[x=ktop, y expr={100*\thisrow{oxford105k}}]  \truncKTop;\leg{Oxford105k};
	\addplot[color={blue},    solid, mark=*,  mark size=1.5, line width=1.0] table[x=ktop, y expr={100*\thisrow{paris106k}}]  \truncKTop;\leg{Paris106k};

\end{axis}
\end{tikzpicture}
}
\end{tabular}
\caption{Retrieval performance (mAP) versus the shortlist size used for affinity matrix truncation.
\label{fig:truncPtg}
\vspace{-10pt}
}
\end{figure}

We now focus on the large scale solutions of Section~\ref{sec:largescale}.

\head{Reduced number of regions.}
Figure~\ref{fig:gmmDiff} shows the impact of reducing the number of regions with Gaussian mixture models. Having as few as 5 descriptors per image already achieves competitive performance, while reducing the online search complexity.
We decrease the number of neighbors $k$ to 50 when GMM reduction is used, as there are now less positive neighbors.
\begin{figure}[b!]
\vspace{-10pt}
\centering
\input{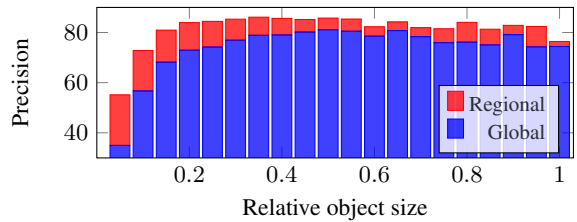}
\small
\begin{tabular}{c}
\extfig{barInstre}{
\begin{tikzpicture}
\begin{axis}[%
	width=0.95\linewidth,
	height=0.43\linewidth,
	ybar stacked,
	reverse legend,
	legend pos=south east,
    legend style={cells={anchor=east}, font =\footnotesize, fill opacity=0.8, row sep=-0.5pt},
    ymax = 90,
    ymin = 30,
    xmin = 0,
    xmax = 1.03,
    xtick={0.2,0.4,...,1},
    ytick={0,20,...,100},
    xlabel={Relative object size},%
    ylabel={Precision},%
 	 x label style={at={(axis description cs:.5,.05)}},
    ylabel style={yshift=-1.5ex}
]
	\addplot[color=blue, fill = blue,fill opacity=0.75,inner ysep=0.5pt, bar width = 7.5] table[x=sz, y expr={100*\thisrow{gPrec}}]  \barGraph;\leg{Global};
	\addplot[color=red, fill = red,fill opacity=0.75,inner ysep=0.5pt, bar width = 7.5] table[x=sz, y expr={100*\thisrow{lPrec}}]  \barGraph;\leg{Regional};
\end{axis}
\end{tikzpicture}
}
\end{tabular}
\caption{Precision of each positive image measured at the position where it was retrieved, averaged over positive images according relative object size. Statistics computed on INSTRE over all queries for global and regional diffusion.
\label{fig:objectsize}
\vspace{-10pt}
}
\end{figure}
\begin{table}
\def \rmac{\scriptsize R\hspace{-0.7pt}-\hspace{-0.7pt}MAC}
\def \aqe{\scriptsize AQE~\cite{CPSIZ07}}
\def \scsm{\scriptsize SCSM~\cite{SLBW14}}
\def \hn{\scriptsize HN~\cite{DGBQG11}}
\def \hqe{\scriptsize HQE}
\def \crow{\scriptsize CroW~\cite{KMO15}}
\def \netvlad{\scriptsize NetVLAD~\cite{AGT+15}}
\def \rmatch{\scriptsize R\hspace{-0.7pt}-\hspace{-0.7pt}match~\cite{RSAC14}}
\def \reg{\scriptsize Regional diffusion}
\def \glob{\scriptsize Global diffusion}
\footnotesize
\begin{center}
    \begin{tabular}{ |@{\sssp}l@{\sssp}|@{\sssp}r@{\sssp}|@{\sssp}c@{\sssp}|@{\sssp}c@{\sssp}|@{\sssp}c@{\sssp}|@{\sssp}c@{\sssp}|@{\sssp}c@{\sssp}|c}
    \hline
      Method                                      &  $m \times d$			& INSTRE    & Oxf5k           & Oxf105k      	    & Par6k       	  & Par106k   	  \\ \hline \hline
    \multicolumn{7}{|c|}{\textbf{ Global descriptors - nearest neighbor search} } \\ \hline \hline
      \crow$^\dagger$                             & 512       				& -         & 68.2            & 63.2              & 79.8            & 71.0          \\
      \rmac~\cite{RTC16}                          & 512       				& 47.7      & 77.7            & 70.1              & 84.1            & 76.8          \\
      \rmac~\cite{GARL16b}                        & 2,048     	      & 62.6			& 83.9    	      & 80.8    	        & 93.8    	      & 89.9     	    \\
      \netvlad$^\dagger$                          & 4,096             & -         & 71.6            & -                 & 79.7            & -             \\ \hline \hline
      \multicolumn{7}{|c|}{\textbf{ Global descriptors - query expansion } }  \\ \hline \hline
      \rmac~\cite{RTC16}+\aqe                     & 512               & 57.3      & 85.4            & 79.7              & 88.4            & 83.5          \\
      \rmac~\cite{RTC16}+\scsm                    & 512               & 60.1      & 85.3            & 80.5              & 89.4            & 84.5          \\
      \rmac~\cite{RTC16}+\hn                      & 512      		 			& 64.7      & 79.9            & -               	& 92.0            & -            	\\
      \glob                                       & 512               & 70.3      & 85.7            & 82.7              & 94.1            & 92.5          \\
      \rmac~\cite{GARL16b}+\aqe                   & 2,048     				& 70.5      & 89.6            & 88.3    	        & 95.3    	      & 92.7     	    \\
      \rmac~\cite{GARL16b}+\scsm                  & 2,048             & 71.4      & 89.1            & 87.3              & 95.4            & 92.5          \\
      \glob                                       & 2,048       			& 80.5      & 87.1            & 87.4          	  & 96.5            & 95.4   	\\ \hline \hline
       \multicolumn{7}{|c|}{\textbf{ Regional descriptors - nearest neighbor search } }  \\ \hline \hline
      \rmatch       			                        & 21$\times$512     & 55.5      & 81.5            & 76.5            	& 86.1            & 79.9          \\
      \rmatch       			                        & 21$\times$2,048   & 71.0      & 88.1            & 85.7            	& 94.9            & 91.3          \\ \hline \hline
       \multicolumn{7}{|c|}{\textbf{ Regional descriptors - query expansion} }  \\ \hline \hline
      \hqe~\cite{TJ14}                            & 2.4k$\times$128   & 74.7      & ~89.4$^\dagger$ & ~84.0$^\dagger$   & ~82.8$^\dagger$ & -             \\

      \rmatch+\aqe                                & 21$\times$512     & 60.4      & 83.6            & 78.6            	& 87.0            & 81.0         	\\
      \reg$^\star$                                & 5$\times$512      & 77.5      & 91.5            & 84.7              & 95.6            & 93.0          \\
      \reg$^\star$                                & 21$\times$512     & 80.0      & 93.2            & 90.3              & 96.5            & 92.6          \\

      \rmatch+\aqe                                & 21$\times$2,048   & 77.1      & 91.0            & 89.6            	& 95.5            & 92.5         	\\
      \reg$^\star$                                & 5$\times$2,048    & 88.4      & 95.0            & 90.0              & 96.4            & \os{95.8}     \\
      \reg$^\star$                                & 21$\times$2,048   & \os{89.6} & \os{95.8}       & \os{94.2}    	    & \os{96.9}   	  & 95.3         	\\  \hline
\end{tabular}
\caption{Performance comparison to the state of the art.
Results from original publications are marked with $^\dagger$, otherwise they are based on our implementation. Our methods are marked with~$^\star$. Points at 512D are extracted with VGG~\cite{RTC16} and at 2048D with ResNet101~\cite{GARL16b}. Regional diffusion with 5 regions uses GMM.
\label{tab:soa}
\vspace{-20pt}
}
\end{center}
\end{table}

\setlength{\fboxsep}{0pt}%
\setlength{\fboxrule}{1.5pt}%
\newcommand{\showIm}[3]{\includegraphics[width=1.4cm,height=1.4cm]{figs/selected/#1_q_#2_#3.jpg}}
\newcommand{\qIm }[2]{\includegraphics[width=1.4cm,height=1.4cm]{figs/selected/{#2}_q_#1_bbx.jpg}}
\newcommand{\pr}[2]{\scriptsize{{\color{red}#1}{\scriptsize$\rightarrow$}{\color{blue}#2}}}
\begin{figure*}[t]
\begin{center} \footnotesize
   \begin{tabular}
   {*{17}{@{\sssp}c@{\sssp}}}

\includegraphics[width=1.25cm,height=1.25cm]{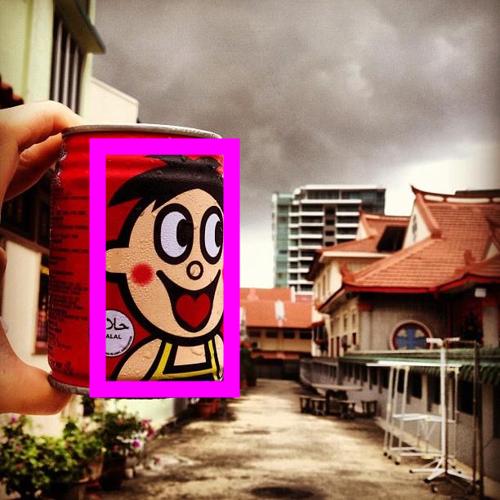} &
\includegraphics[width=1.25cm,height=1.25cm]{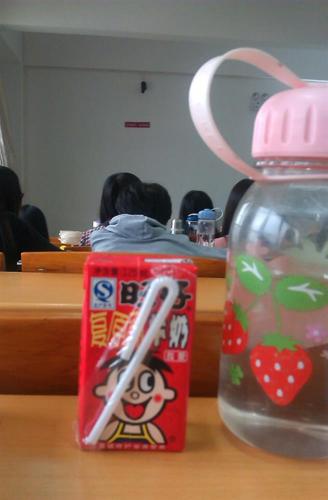} &
\includegraphics[width=1.25cm,height=1.25cm]{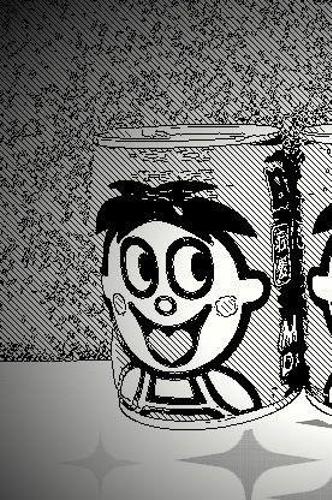} &
\includegraphics[width=1.25cm,height=1.25cm]{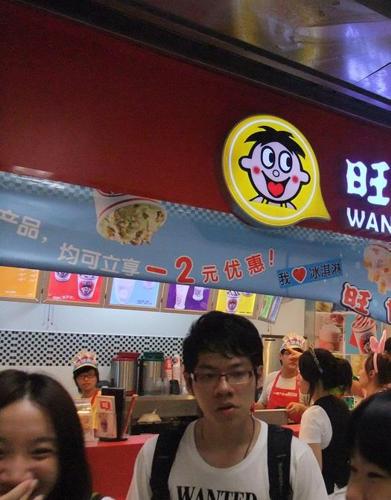} &
\includegraphics[width=1.25cm,height=1.25cm]{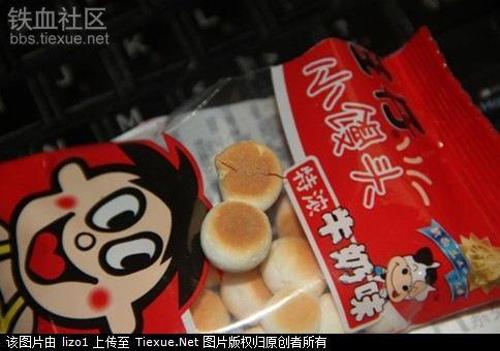} &
\includegraphics[width=1.25cm,height=1.25cm]{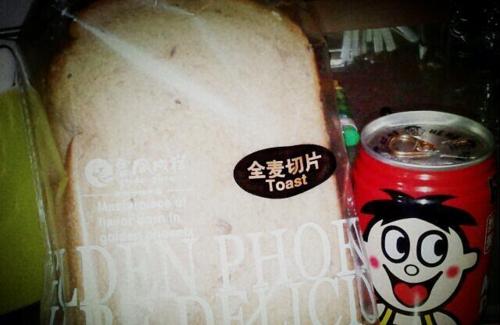} &
\includegraphics[width=1.25cm,height=1.25cm]{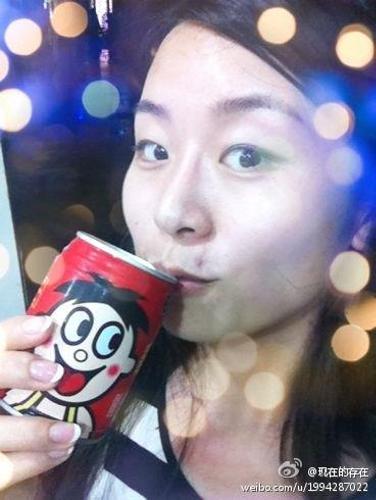} &
\includegraphics[width=1.25cm,height=1.25cm]{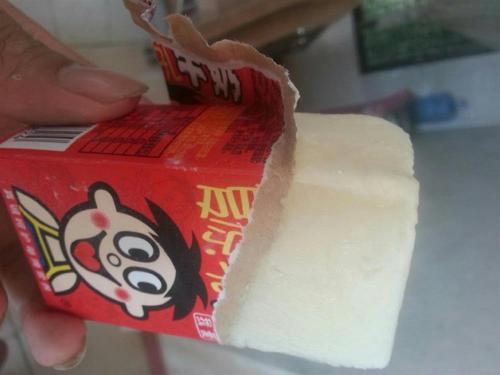} &
\includegraphics[width=1.25cm,height=1.25cm]{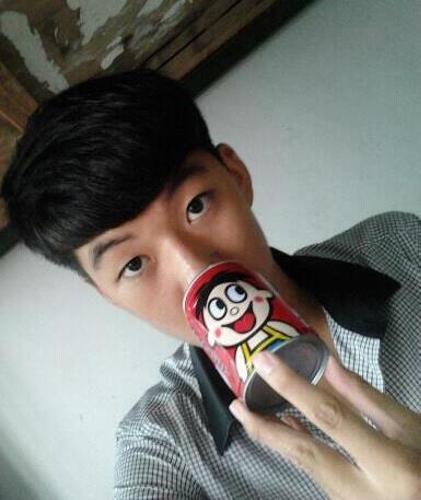} &
\includegraphics[width=1.25cm,height=1.25cm]{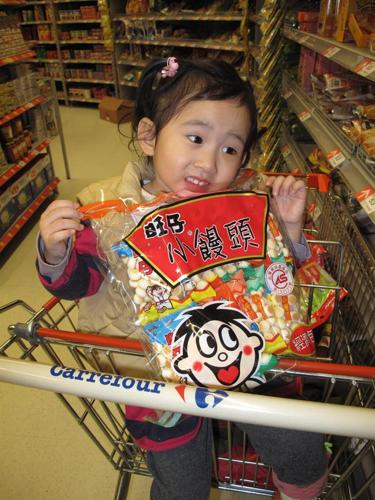} &
\includegraphics[width=1.25cm,height=1.25cm]{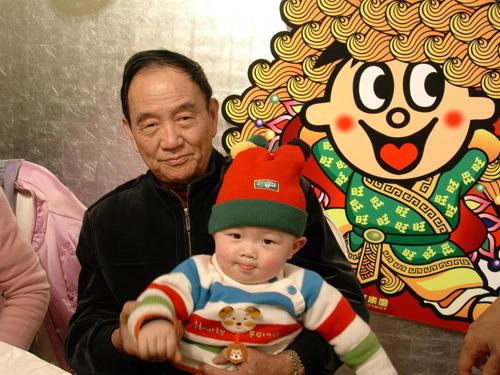} &
\includegraphics[width=1.25cm,height=1.25cm]{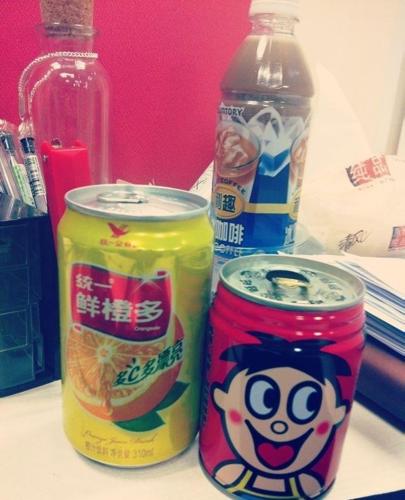} &
\includegraphics[width=1.25cm,height=1.25cm]{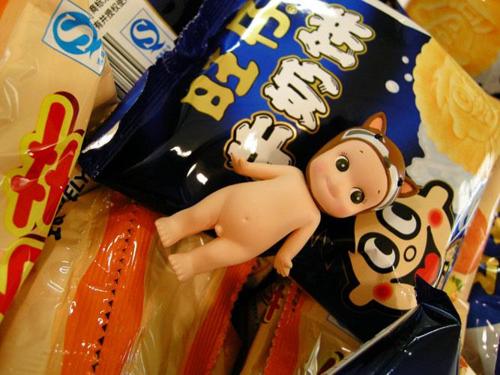} 
\\[-2pt]
(AP: \pr{43.1}{84.9})&
\pr{0.5}{100} &
\pr{0.6}{100} &
\pr{1.8}{100} &
\pr{0.6}{98.7} &
\pr{2.6}{100} &
\pr{2.6}{100} &
\pr{0.4}{97.7} &
\pr{1.6}{98.8} &
\pr{3.3}{100}&
\pr{0.4}{96.7}&
\pr{4.5}{100}&
\pr{3.7}{98.8}
\\[3pt]
\includegraphics[width=1.25cm,height=1.25cm]{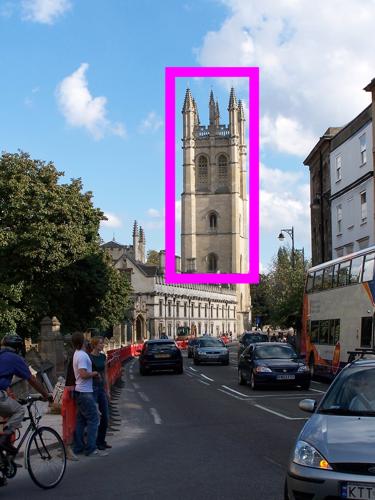} &
\includegraphics[width=1.25cm,height=1.25cm]{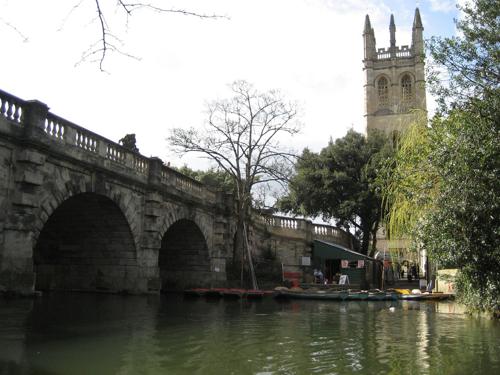} &
\includegraphics[width=1.25cm,height=1.25cm]{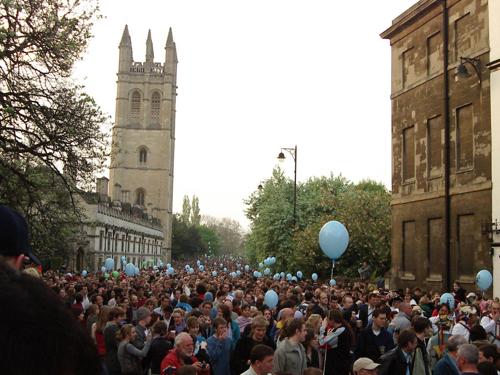} &
\includegraphics[width=1.25cm,height=1.25cm]{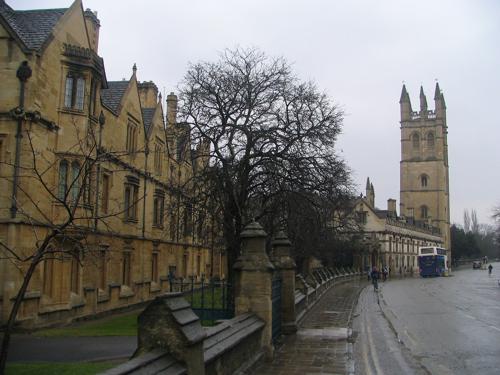} &
\includegraphics[width=1.25cm,height=1.25cm]{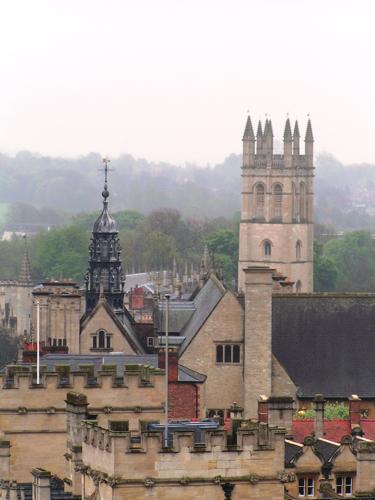} &
\includegraphics[width=1.25cm,height=1.25cm]{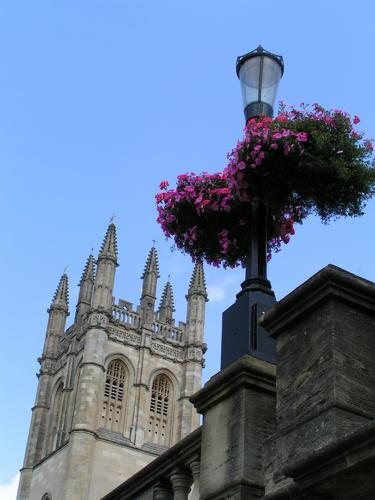} &
\includegraphics[width=1.25cm,height=1.25cm]{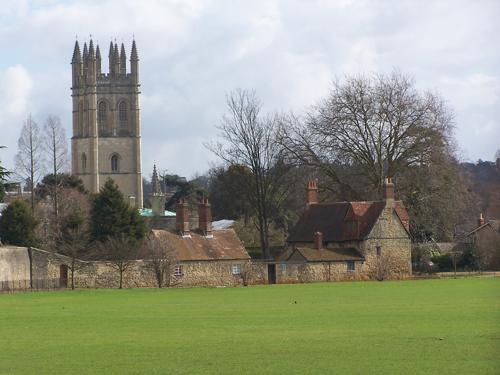} &
\includegraphics[width=1.25cm,height=1.25cm]{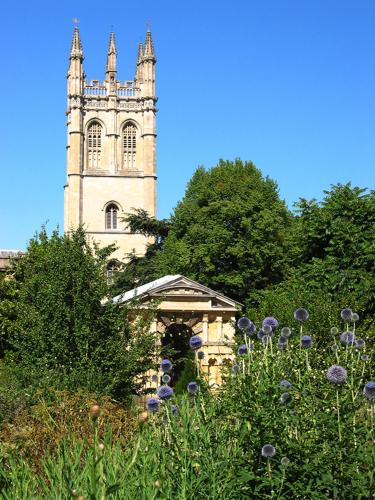} &
\includegraphics[width=1.25cm,height=1.25cm]{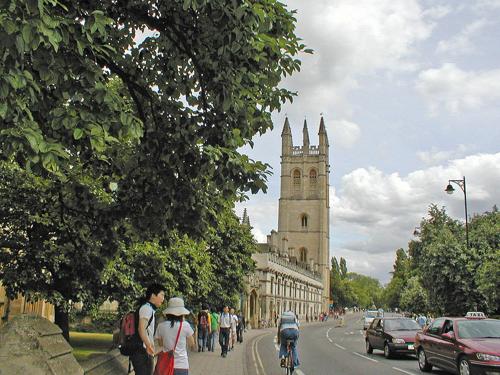} &
\includegraphics[width=1.25cm,height=1.25cm]{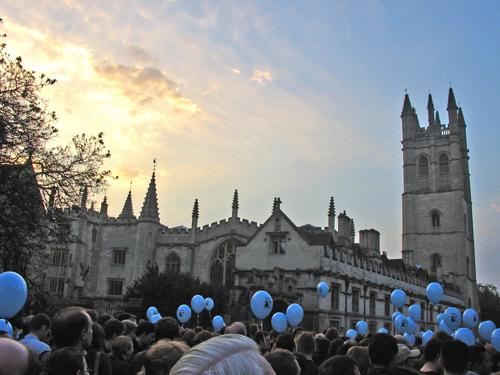} &
\includegraphics[width=1.25cm,height=1.25cm]{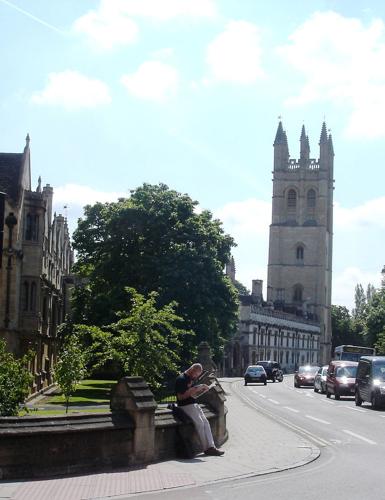} &
\includegraphics[width=1.25cm,height=1.25cm]{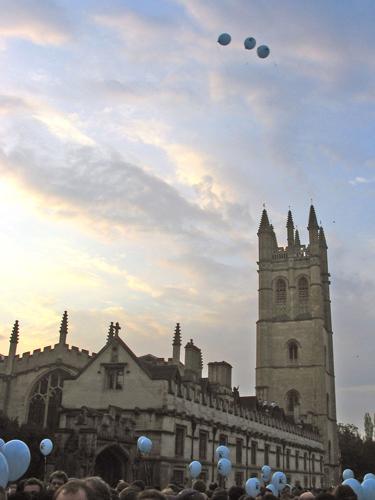} &
\includegraphics[width=1.25cm,height=1.25cm]{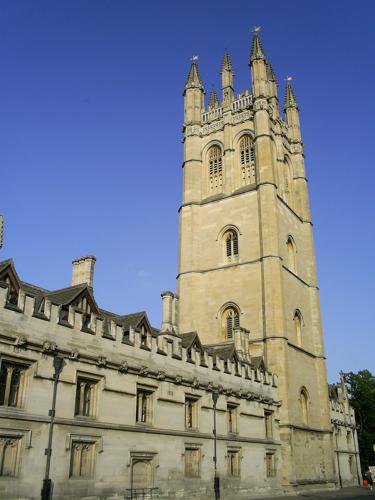} 
\\[-2pt]
(AP: \pr{24.0}{89.9}) &
\pr{3.1}{100}&
\pr{4.2}{100}&
\pr{4.3}{100}&
\pr{4.7}{100}&
\pr{5.9}{100}&
\pr{5.9}{100}&
\pr{6.4}{100}&
\pr{7.0}{100}&
\pr{7.0}{100}&
\pr{7.0}{100}&
\pr{7.1}{100}&
\pr{7.3}{100}
\\[3pt]
\includegraphics[width=1.25cm,height=1.25cm]{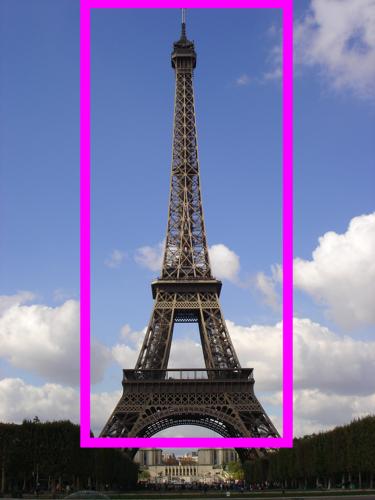} &
\includegraphics[width=1.25cm,height=1.25cm]{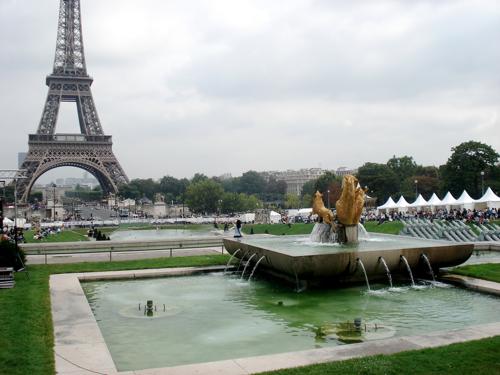} &
\includegraphics[width=1.25cm,height=1.25cm]{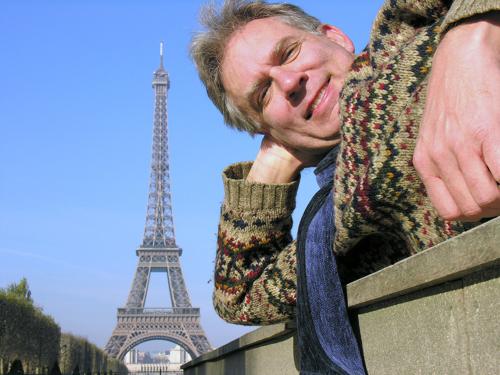} &
\includegraphics[width=1.25cm,height=1.25cm]{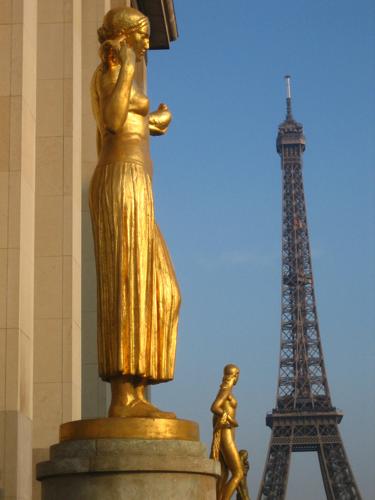} &
\includegraphics[width=1.25cm,height=1.25cm]{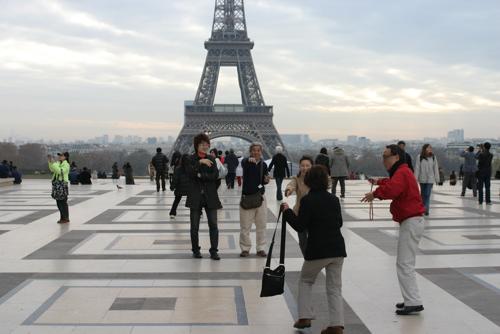} &
\includegraphics[width=1.25cm,height=1.25cm]{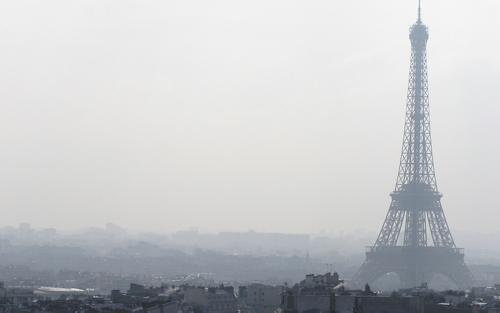} &
\includegraphics[width=1.25cm,height=1.25cm]{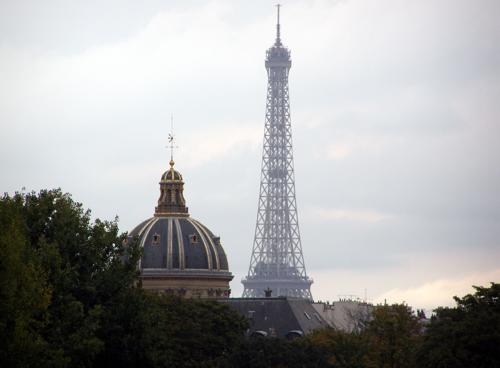} &
\includegraphics[width=1.25cm,height=1.25cm]{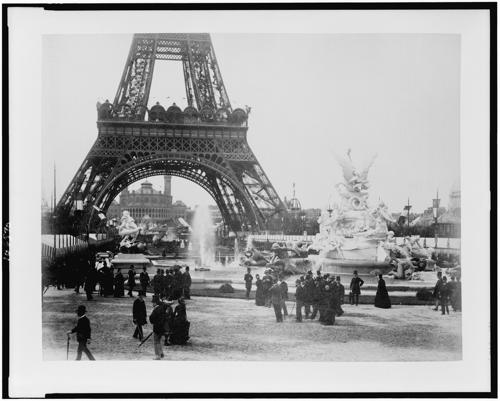} &
\includegraphics[width=1.25cm,height=1.25cm]{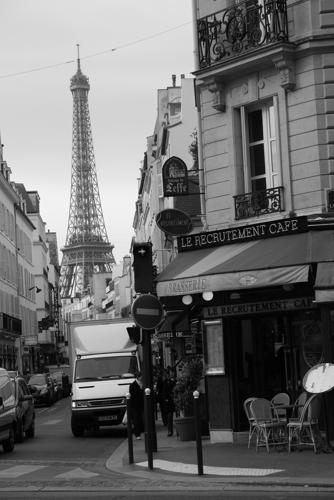} &
\includegraphics[width=1.25cm,height=1.25cm]{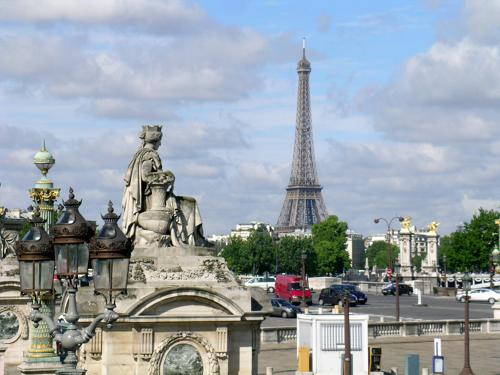} &
\includegraphics[width=1.25cm,height=1.25cm]{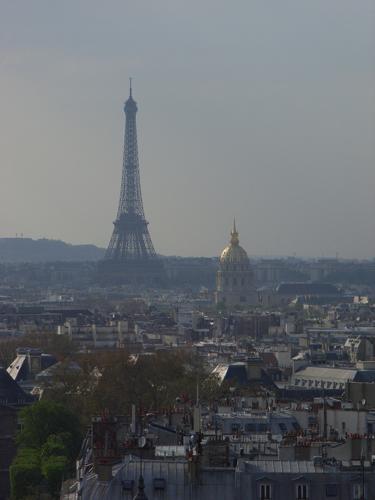} &
\includegraphics[width=1.25cm,height=1.25cm]{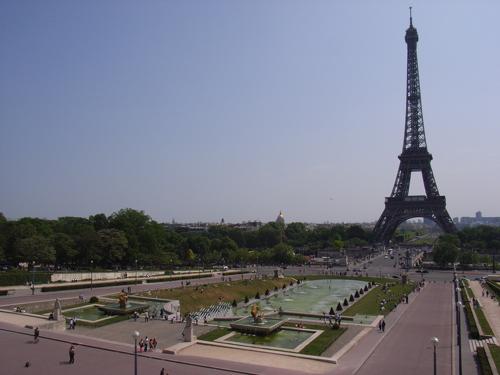} &
\includegraphics[width=1.25cm,height=1.25cm]{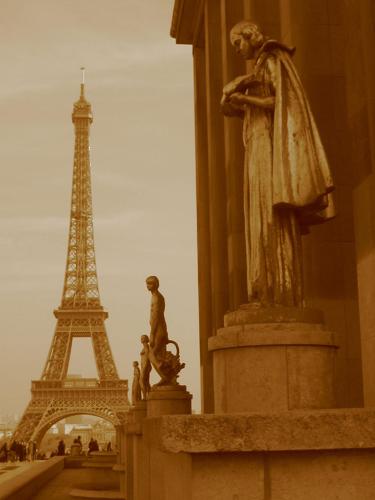} 
\\[-2pt]
(AP: \pr{56.5}{94.3}) &
\pr{8.2}{94.5}&
\pr{12.9}{91.5}&
\pr{18.8}{91.6}&
\pr{14.7}{85.4}&
\pr{13.3}{83.6}&
\pr{15.9}{86.1}&
\pr{15.9}{84.8}&
\pr{10.9}{79.5}&
\pr{13.7}{82.4}&
\pr{17.0}{84.1}&
\pr{17.8}{84.4}&
\pr{26.5}{93.0}
\\[3pt]

  \end{tabular}
    \caption{Query examples from INSTRE, Oxford, and Paris datasets and retrieved images ranked by decreasing order of ranking difference between global and regional diffusion. We measure precision at the position where each image is retrieved and report this under each image for {\color{red}global} and {\color{blue}regional} diffusion. Average Precision (AP) is reported per query for the two methods.
    \label{fig:examples}}
\end{center}
\end{figure*}

\head{Affinity matrix with Dong's algorithm~\cite{DCL11}.}
We compare the exhaustive construction of matrix $A$ to Dong's efficient $k$-NN graph algorithm~\cite{DCL11}. Exhaustive search for Oxford105k composed of 2.2M regions takes 96 hours on a machine with a 12-core Intel Xeon 2.30GHz CPU. The approximate graph only takes 45 minutes and affects the final retrieval performance only slightly. It achieves 91.6 mAP on Oxford105k and 94.6 on Paris106k, while the exhaustive construction yields 92.5 and 95.2 respectively.

\head{Truncation} is a means to handle large scale datasets, \ie more than 100k images.
Regional diffusion on the full dataset takes 13.9$s$ for Oxford105k,
 which is not practical. We therefore rank images according to the aggregated regional descriptors, which is equivalent to the R-MAC representation~\cite{TSJ15}, and then perform diffusion on a short-list. Figure~\ref{fig:truncPtg} reports results with truncation. The performance of the full database diffusion is nearly attained by re-ranking less than 10\% of the database.
The entire truncation  and diffusion process on Oxford105k takes 1s, with truncation and re-normalization taking only a small part of it.
In the following, search on Oxford105k and Paris105k is performed by truncating the top 10k images.
This choice results in an affinity matrix $A$ of around 200k regions.
When GMM reduction is used, our short-list size is chosen so that $A$ has 2M regions too, keeping re-ranking complexity fixed.

Our approach is scalable thanks to truncation: the shortlist length is fixed and so is the re-ranking time, regardless of the database size and the dimensionality of the descriptors.
Although this shortlist contains a small fraction of the database, its significantly outperforms the baseline.

\head{Small objects.} We present quantitative and qualitative results revealing that images benefit from our method mainly when the depicted object is small and the scene is cluttered. Figure~\ref{fig:examples} shows that the retrieved images with the highest increase of precision of regional compared to global diffusion contain small objects that the latter cannot see.
Since the bounding boxes are available for all images of INSTRE, we quantitatively measure precision for all positive images: Figure~\ref{fig:objectsize} shows that the highest improvement indeed comes for objects with small relative size.

\subsection{Comparison to other methods}
\label{sec:comparison}
We compare with the state-of-the-art approaches with global or regional representation, with or without query expansion.
Table~\ref{tab:soa} summarizes the results.
We implement three methods typically combined with BoW, namely Average Query Expansion (AQE)~\cite{CPSIZ07}, Spatially Constrained Similarity Measure (SCSM)~\cite{SLBW14} and Hello Neighbor (HN)~\cite{DGBQG11}.
AQE is also effective with CNN global representation~\cite{TSJ15,KMO15,GARL16}.
A baseline for the regional scenario is R-match~\cite{RSAC14}.
We additionally extend AQE to regional representation\footnote{AQE has not been proposed in a regional scenario. We extend it as competitive baseline derived from prior work.} combined with the similarity used in R-match.
Hamming Query Expansion\footnote{We evaluated HQE on INSTRE for the purposes of this work.} (HQE)~\cite{TJ14} is the only method not using CNNs, but local descriptors.

Regional diffusion significantly outperforms all other methods in all datasets.
Global diffusion performs well on Paris because query objects almost fully cover the image in most of the database entries.
This does not hold on INSTRE, which contains a lot of small objects. The improvements of regional diffusion are in this case much larger.
%
%

\vspace{-5pt}
\section{Conclusion}
\label{sec:conclusion}
\vspace{-5pt}
We propose a retrieval approach capturing distinct manifolds in the description space
at no additional cost compared to a single query.
We experimentally show that it significantly improves retrieval of small objects and cluttered scenes. The conclusion is that as few as 5-10 regional CNN descriptors can convey important information on small objects while thousands of conventional local descriptors are typically needed. Thus, a regional affinity matrix becomes possible. Regional diffusion was not possible before.
In contrast to prior work, we use the closed form solution of the diffusion iteration, obtained by the conjugate gradient method. Combined with our contributions on space efficiency, this achieves large scale search at reasonable query times.
Using recent CNN architectures, we achieve state-of-the-art and near optimal performance on two popular benchmarks
and
a recent more challenging dataset.

\head{Acknowledgments}
The authors were supported by the MSMT LL1303 ERC-CZ grant. The Tesla K40 used for this research was donated by the NVIDIA Corporation.

{\small
\bibliographystyle{ieee}
\bibliography{egbib}
}
\flushend
\end{document}